\documentclass[runningheads]{llncs}
\usepackage{graphicx}
\usepackage{hyperref}
\usepackage{url}            % simple URL typesetting
\usepackage{amsfonts}       % blackboard math symbols
\usepackage{nicefrac}       % compact symbols for 1/2, etc.
\usepackage{microtype}      % microtypography
\usepackage{graphicx}
\usepackage{amsmath,amssymb}
\usepackage{bm}
\usepackage{array,booktabs,ragged2e}
\usepackage{cleveref}
\usepackage{subfig}

\begin{document}
\title{Learning to Adapt Dynamic Clinical Event Sequences with Residual Mixture of Experts
}
\titlerunning{Learning to Adapt Clinical Sequences with Residual Mixture of Experts}
\vspace{-9mm}

\author{Jeong Min Lee\orcidID{0000-0001-8630-0546} \and Milos Hauskrecht\orcidID{0000-0002-7818-0633}}

\institute{Department of Computer Science, University of Pittsburgh, Pittsburgh, PA, USA
\email{\{jlee, milos\}@cs.pitt.edu}}
\authorrunning{J. Lee and M. Hauskrecht}
% \institute{}
%
\maketitle              % typeset the header of the contribution
\begin{abstract}
\vspace{-4mm}
Clinical event sequences in Electronic Health Records (EHRs) record detailed information about the patient condition and patient care as they occur in time. Recent years have witnessed increased interest of machine learning community in developing machine learning models solving different types of problems defined upon information in EHRs. 
More recently, neural sequential models, such as RNN and LSTM, became popular and widely applied models for representing patient sequence data and for predicting future events or outcomes based on such data. However, a single neural sequential model may not properly represent complex dynamics of all patients and the differences in their behaviors. In this work, we aim to alleviate this limitation by refining a one-fits-all model using a Mixture-of-Experts (MoE) architecture. The architecture consists of multiple (expert) RNN models covering patient sub-populations and refining the predictions of the base model. That is, instead of training expert RNN models from scratch we define them on the residual signal that attempts to model the differences from the population-wide model. The heterogeneity of various patient sequences is modeled through multiple experts that consist of RNN. Particularly, instead of directly training MoE from scratch, we augment MoE based on the prediction signal from pretrained base GRU model. With this way, the mixture of experts can provide flexible adaptation to the (limited) predictive power of the single base RNN model. 
We experiment with the newly proposed model on real-world EHRs data and the multivariate clinical event prediction task. We implement RNN using Gated Recurrent Units (GRU). We show 4.1\% gain on AUPRC statistics compared to a single RNN prediction.

\end{abstract}

\section{Introduction}
Clinical event sequences in Electronic Health Records (EHRs) record detailed information about the patient condition and patient care as they occur in time. In recent years, we have witnessed increased interest of machine learning community in developing machine learning models and solutions to different types of problems defined upon information in EHRs. Examples of the problems are automatic patient diagnosis \cite{malakouti2019hierarchical,malakouti2019predicting}, mortality prediction \cite{yu2020monitoring}, detection of patient management errors \cite{hauskrecht_outlier_2013,hauskrecht2016outlier}, length of stay predictions \cite{miotto2016deep},
or sepsis prediction \cite{henry2015targeted}.
Interestingly, the overwhelming majority of the proposed solutions in recent years are based on applications of modern autoregressive models based on Recurrent neural networks (RNN), such as LSTM or GRU, and their variants. However, the problem with many of these works is that they aim to learn one autoregressive model to fit all patients and their records. This may limit the ability of the models to represent well the dynamics of heterogeneous patient subpopulations their trends, and outcomes. 

The goal of this paper is to study ways of enhancing the one-fits-all RNN model solution with additional RNN models to better fit the differences in patient behaviors and patient subpopulations. We study this solution in the context of multivariate event prediction problem where our goal is to predict, as accurately as possible, the future occurrence of a wide range of events recorded in EHRs. Such a prediction task can be used for defining general predictive patient state representation that can be used, for example, to define similarity among patients or for predicting patient outcomes. 

To represent different patient subpopulations and their behaviors we explore and experiment with the mixture of experts architecture \cite{jacobs1991adaptive}. However, instead of training multiple RNN models from scratch, we first train the all-population model and train the mixture on the residual signals, such that the models augment the output of the all population model. Hence we name our model Residual Mixture of Experts (R-MoE). The rationale for such a design is that residual subpopulation models may be much simpler and hence easier to learn. Moreover, this design simplifies the subpopulation model switching, that is, more subpopulation models can be used to make predictions for the same patient at different times. 

R-MoE aims to provide flexible adaptation to the (limited) predictive power of the population model.
We demonstrate the effectiveness of R-MoE on the multivariate clinical sequence prediction task for real-world patient data from MIMIC-3 Database \cite{johnson2016mimic}. The experiments with R-MoE model show 4.1\% gain on AUPRC compared to a single GRU-based prediction. 
For the reproducibility, the code, trained models, and data processing scripts are available on this link: \url{https://github.com/leej35/residual-moe}

\section{Related Work}

\subsubsection{Subpopulation and Adaptive Models.} The goal of the adaptive models is to address the drift of distribution in data. In clinical and biomedical research areas, this is a particularly important issue due to the heterogeneity of overall patient population and its subpopulations. For instance, when a model that is trained on the overall patient population is used for a target patient that belongs to a certain subpopulation, there is no guarantee that the model would perform well. 

One traditional approach used to address the problem is to divide (cluster) the patients into subgroups using a small set of patients' characteristics (features) and train many different subpopulation models for these groups. The subpopulation model for the target patient is chosen by matching the corresponding group features \cite{huang2015medical,huang2013similarity}. Another related approach is to flexibly identify the subpopulation group that closely matches the target patient through similarity measures and build the subpopulation model for the target patient from this patient group \cite{fojo2017precision,visweswaran2005instance}.  
An alternative way is to adapt the parameters of the population models to fit the target patient. In the NLP area, non-parametric memory components are used to build adaptive models that sequentially update the model's parameters \cite{grave_unbounded_nodate,krause2018dynamic}. 
In the clinical domain, simpler residual models that learn the difference (residuals) between the predictions made by population models and the desired outcomes are learned for continuous-valued clinical time series and achieve better forecasting performance \cite{liu2016learning_a}.  

\vspace{-3mm}
\subsubsection{Neural Clinical Sequence Models.}
Neural-based models have become more popular in recent years for modeling clinical sequence data. Especially, they have advantages such as flexibility in modeling latent structures and capability in learning complex patient state dynamics. 
More specifically, word embedding methods (e.g., CBOW, Skip-gram) are used to obtain compact representation of clinical concepts \cite{choi2016medical,choi2016learning} and predictive patient state representations \cite{choi2017using}.
For autoregressive clinical sequence modeling, neural temporal models (e.g., RNN, GRU) and attention mechanism are used to learn patient state dynamics \cite{lee2019context,lee2021neural}, predict future states progression \cite{lee2020multi}, and predict clinical variables such as
diagnosis codes \cite{malakouti2019hierarchical,malakouti2019predicting}, readmission of chronic diseases \cite{Nguyen2017FindingAS}, medication prescriptions \cite{Bajor2017PredictingMF}, ICU mortality risk \cite{yu2020monitoring}, disease progression of diabetes and mental health \cite{pham2017predicting}, and multivariate future clinical event occurrences \cite{lee_clinical_2020,lee2021modeling,liu2019nonparametric}.

\section{Methodology}

\subsection{Neural Event Sequence Prediction} \label{sec:3.1}
In this work, our goal is to predict the occurrences and non-occurrences of future clinical (target) events $\bm{y}'_{t+1}$ for a patient given the patient's past clinical event occurrences $\bm{H}_t$. Specifically, we assume that the patient's clinical event history is in a sequence of multivariate input event vectors $\bm{H}_t = \{\bm{y}_{1},\dotsc,\bm{y}_{t}\}$ where each vector $\bm{y}_{i}$ is a binary $\{0,1\}$ vector, one dimension per an event type. The input vectors are of dimension $|E|$ where $E$ are different event types in clinical sequences. The target vector is of dimension $|E'|$, where $E'\subset E$ are events we are interested in predicting. We aim to build a predictive model $\delta$ that can predict $\hat{\bm{y}}'_{t+1}$ at any time $t$ given the history $\bm{H}_t$. 

One way to build $\delta$ is to use neural sequence models such as RNN and LSTM. In this work, we use Gated Recurrent Units (GRU) \cite{cho2014learning} to build a base prediction model $\delta_{base}$ with input embedding matrix $\bm{W}_{emb} \in \mathbb{R}^{|E| \times \epsilon}$, output projection matrix $\bm{W}_{o} \in \mathbb{R}^{d \times |E'|}$, bias vector $\bm{b}_{o} \in \mathbb{R}^{|E'|}$, and a sigmoid (logit) activation function $\sigma$. At any time step $t$, we update hidden state $\bm{h}_t$ and predict target events in next time step $\bm{\hat{y}}'_{t+1}$: 

\[ \begin{array}{lll}%
\bm{v}_t = \bm{W}_{emb} \cdot \bm{y}_t &\qquad 
\bm{h}_{t} = \text{GRU}(\bm{h}_{t-1}, \bm{v}_t) &\qquad 
\bm{\hat{y}}'_{t+1} = \sigma(\bm{W}_{o} \cdot \bm{h}_{t} + \bm{b}_{o})\\
\end{array}\]%

All parameters of $\delta_{base}$ ($\bm{W}_{emb}, \bm{W}_{o}, \bm{b}_{o}$ and GRU) are learned through stochastic gradient descent (SGD) algorithm with the binary cross entropy loss function.

This GRU-based neural sequence model has a number of benefits for modeling complex high-dimensional clinical event time-series: First, we can obtain a compact real-valued representation of high-dimensional binary input vector $\bm{y}$ through low-dimensional embedding with $\bm{W}_{emb}$. Second, we can model complex dynamics of patient state sequences through GRU which can model non-linearities of the sequences. Third, complex input-output associations of the patient state sequences can be learned through a flexible SGD-based end-to-end learning framework. 

Nonetheless, the neural approach cannot address one important peculiarity of the patient state sequence: the heterogeneity of patient sequences across patient populations. Typically, clinical event sequences in EHRs are generated from a pool of diverse patients where each patient has different types of clinical complications, medication regimes, or observed sequence dynamics. While the average behavior of clinical event sequences can be captured well by a single neural sequence model, the model may fail to represent the detailed dynamics of heterogeneous clinical event sequences for individual patients.  

\begin{figure}[t]
\begin{center}

\centerline{\includegraphics[scale=0.42]{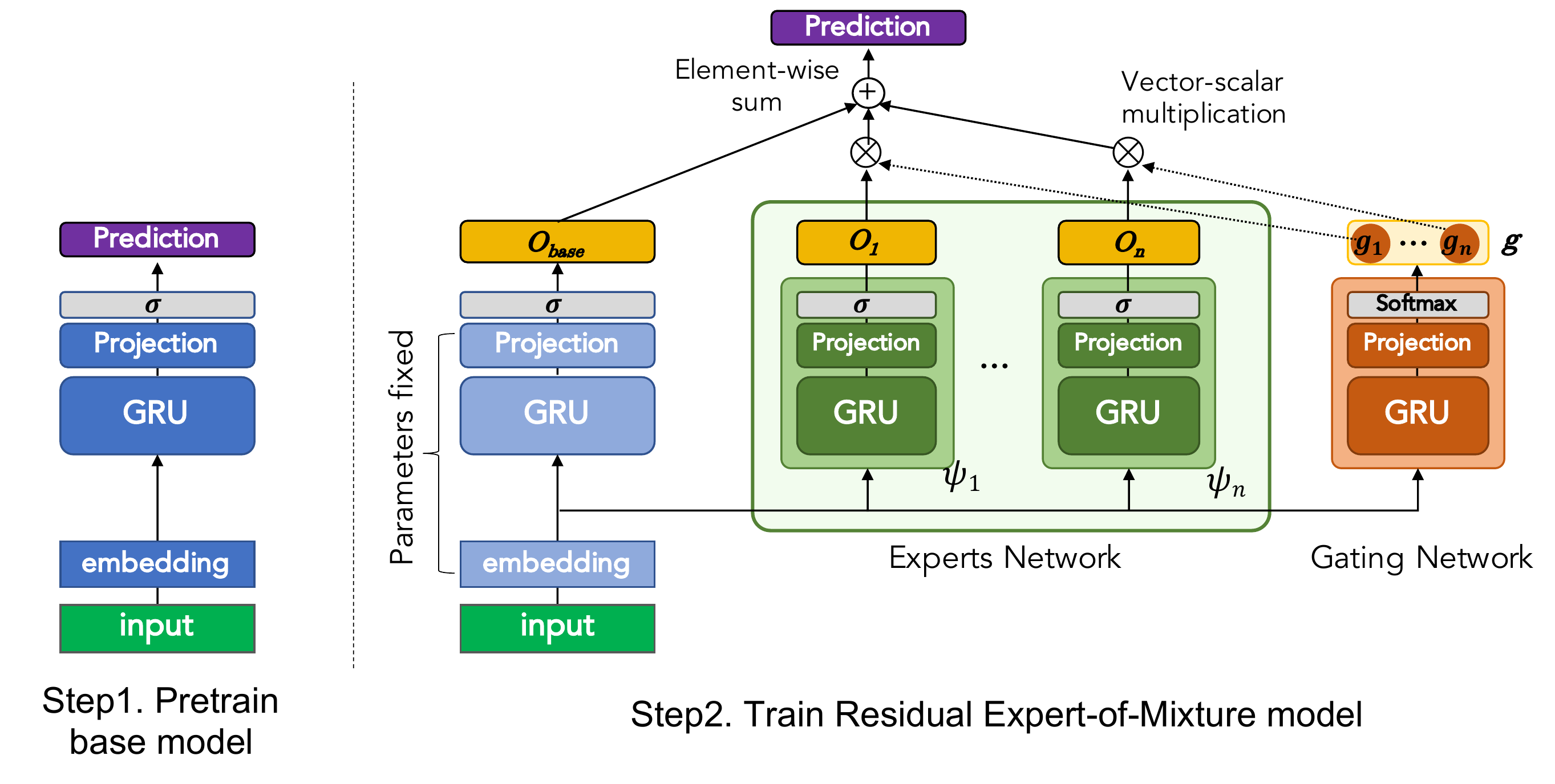}}

\caption{Architecture of R-MoE model. First we train base model $\delta_{base}$. Then, we fix parameters of $\delta_{base}$ and train parameters of Mixture-of-Experts consists of Experts network and Gating network with the combined prediction of $\delta_{base}$ and the MoE. With this way, MoE can learn to adapt the residual of $\delta_{base}$.}
\label{figure:overall_architecture}
\end{center}
\vspace{-6mm}
\end{figure}

\subsection{Residual Mixture-of-Experts}

In this work, we address the heterogeneity issue of the neural sequence model by specializing it with a novel learning mechanism based on Mixture-of-Experts (MoE) architecture. The dynamics of heterogeneous patient state sequences can be modeled through a number of experts; each consists of GRU which is capable of modeling non-linearities and temporal dependencies. Particularly, in this work, instead of simply replacing the GRU model $\delta_{base}$ with MoE, we \textit{augment} $\delta_{base}$ with MoE. The key idea is to specialize the Mixture-of-Experts to learn the residual that $\delta_{base}$ cannot capture. As shown in Figure \ref{figure:overall_architecture}, the proposed model R-MoE consists of $\delta_{base}$ module and Mixture-of-Experts module. 

The Mixture-of-Experts module consists of $n$ experts $\psi_{1},\dotsc,\psi_{n}$ and a gating network $G$ which outputs a $n$-dimensional vector $\bm{g}$. The output $\bm{o}_{moe}$ of the MoE module can be written as follows:
\begin{equation}
    \label{eq_o_moe}
    \bm{o}_{moe} = \sum_{i=1}^{n} \bm{g}_{[i]}(\bm{v}_t) \cdot \psi_{i}(\bm{v}_t)  
\end{equation}

Each expert $\psi_{i}$ consists of GRU with its hidden state dimension $d'$, output projection matrix $\bm{W}_{o}^{i} \in \mathbb{R}^{d' \times |E'|}$, and a bias vector $\bm{b}_{o}^{i} \in \mathbb{R}^{|E'|}$. Given an input in low-dimensional representation $\bm{v}_t$, an expert $\psi_{i}$ outputs $\bm{o}^{i}$:
\[ \begin{array}{lll}%
\bm{h}_{t}^{i} = \text{GRU}^{i}(\bm{h}^{i}_{t-1}, \bm{v}_t) &\qquad 
\bm{o}^{i} = \sigma(\bm{W}_{o}^{i} \cdot \bm{h}_{t}^{i} + \bm{b}_{o}^{i}) &\qquad 
i \in 1,\dotsc,n\\
\end{array}\]%

The gating network $G$ have the same input and a similar architecture, except that its output $\bm{g}$'s dimension is $n$ and it is through $Softmax$ function. $\bm{g}_{[i]}$ in Eq. \ref{eq_o_moe} represents $i$ value in the vector $\bm{g}$ .
\[ \begin{array}{ll}%
\bm{h}_{t}^{g} = \text{GRU}^{g}(\bm{h}^{g}_{t-1}, \bm{v}_t) &\qquad 
\bm{g} = Softmax(\bm{W}_{o}^{g} \cdot \bm{h}_{t}^{g} + \bm{b}_{o}^{g})\\
\end{array}\]%

The final prediction $\bm{\hat{y}}'_{t+1}$ is generated by summing outputs of the two modules $\bm{o}_{base} = \delta_{base}(\bm{H}_t)$ and $\bm{o}_{moe}$:
\begin{equation}
    \label{rmoe_pred}
    \bm{\hat{y}}'_{t+1} = \bm{o}_{base} + \bm{o}_{moe}
\end{equation}

To properly specialize the Mixture-of-Experts on the residual, we train the two modules as follows: First, we train $\delta_{base}$ module, and parameters of $\delta_{base}$ are fixed after the train. Then, we train R-MoE module with the binary cross entropy loss computed with the final prediction in Eq. \ref{rmoe_pred}.
With this way, R-MoE can learn to adapt the residual, which the base GRU cannot properly model. R-MoE provides flexible adaptation to the (limited) predictive power of the base GRU model.  

\section{Experimental Evaluation}

In this section, we evaluate the performance of R-MoE model on the real-world EHRs data in MIMIC-3 Database \cite{johnson2016mimic} and compare it with alternative baselines. 

\subsection{Experiment Setup} \label{sec:exp-setup}

\subsubsection{Clinical Sequence Generation.}  From MIMIC-3 database, we extract 5137 patients using the following criteria: (1) length of stay of the admission is between 48 and 480 hours, (2) patient's age at admission is between 18 and 99, and (3) clinical records are stored in Meta Vision system, one of the systems used to create MIMIC-3. We split all patients into train and test sets using 80:20 split ratio. 
From the extracted records, we generate multivariate clinical event time series with a sliding-window method. We segment all patient event time series with a time-window $W$=$24$. All events that occurred in a time-window are aggregated into a binary vector $\bm{y}_{i} \in \{0,1\}^{|E|}$ where $i$ denotes a time-step of the window and $E$ is a set of event types. At any point of time $t$, a sequence of vectors created from previous time-windows $\bm{y}_{1},\dotsc,\bm{y}_{t-1}$ defines an (input) sequence. A vector representing target events in the next time-window defines the prediction target $\bm{y'}_{t}$. In our work, a set of target events $E'$ that we are interested in predicting is a subset of input event $E$. 

\vspace{-2mm}
\subsubsection{Feature Extraction.} For our study, we use clinical events in medication administration, lab results, procedures, and physiological results categories in MIMIC-3 database. For the first three categories, we remove events that were observed in less than 500 different patients. For physiological events, we select 16 important event types with the help of a critical care physician. Lab test results and physiological measurements with continuous values are discretized to high, normal, and low values based on normal ranges compiled by clinical experts. In terms of prediction target events $E'$, we only consider and represent events corresponding to occurrences of such events, and we do not predict their normal or abnormal values. This process results in 65 medications, 44 procedures, 155 lab tests, and 84 physiological events as prediction targets for the total target vector size of 348. The input vectors are of size 449.

\vspace{-2mm}
\subsubsection{Baseline Models.} 
We compare R-MoE with multiple baseline models that are able to predict events for multivariate clinical event time-series given their previous history. The baselines are: 
\begin{itemize}
\item \textbf{Base GRU model (GRU)}: GRU-based event time-series modeling described in \Cref{sec:3.1}. ($\lambda$=1e-05)
\item \textbf{REverse-Time AttenTioN (RETAIN)}: RETAIN is a representative work on using attention mechanism to summarize clinical event sequences, proposed by Choi et al. \cite{choi2016retain}. It uses two attention mechanisms to comprehend the history of GRU-based hidden states in reverse-time order. For multi-label output, we use a sigmoid function at the output layer.  ($\lambda$=1e-05)
\item \textbf{Logistic regression based on Convolutional Neural Network (CNN)}: This model uses CNN to build predictive features summarizing the event history of patients. Following Nguyen et al. \cite{nguyen2016mathtt}, we implement this CNN-based model with a 1-dimensional convolution kernel followed by ReLU activation and max-pooling operation. To give more flexibility to the convolution operation, we use multiple kernels with different sizes (2,4,8) and features from these kernels are merged at a fully-connected (FC) layer. ($\lambda$=1e-05)
\item \textbf{Logistic Regression based on the Full history (LR)}: This model aggregates all event occurrences from the complete past event sequence and represents them as a binary vector. The vector is then projected to the prediction with an FC layer followed by an element-wise sigmoid function.
\end{itemize}

\subsubsection{Model Parameters.} 
We use embedding dimension $\epsilon$=$64$, hidden state dimension $d$=$512$ for base GRU model and RETAIN. Hidden states dimension $d'$ for each GRU in R-MoE is determined by the internal cross-validation set (range: 32, 64, 128, 256, 512). The number of experts for R-MoE is also determined by internal cross-validation set (range:1, 5, 10, 20, 50, 100). For the SGD optimizer, we use Adam \cite{kingma2014adam}. For learning rate of GRU, RETAIN, CNN, and LR we use $0.005$ and for R-MoE we use $0.0005$. To prevent over-fitting, we use L2 weight decay regularization during the training of all models and weight $\lambda$ is determined by the internal cross-validation set. Range of $\lambda$ for GRU, RETAIN, CNN, and LR is set as (1e-04, 1e-05, 1e-06, 1e-07). For R-MoE, after observing it requires a much larger $\lambda$, we set the range of $\lambda$ for R-MoE as (0.75, 1.0, 1.25, 1.5). We also use early stopping to prevent over-fitting. That is, we stop the training when the internal validation set's loss does not improve during the last $K$ epochs ($K$=$5$). 

\subsubsection{Evaluation Metric.} We use the area under the precision-recall curve (AUPRC) as the evaluation metric. AUPRC is known for presenting a more accurate assessment of the performance of models for a highly imbalanced dataset \cite{saito2015precision}.

\subsection{Results}

Table \ref{table:overall_results} summarizes the performance of R-MoE and baseline models. The results show that R-MoE clearly outperforms all baseline models. More specifically, compared to GRU, the best-performing baseline model, our model shows 4.1\% improvement. Compared to averaged AUPRC of all baseline models, our model shows 8.4\% gain. 

\begin{table}[h]
\centering
\setlength{\tabcolsep}{1em}
\begin{tabular}{lrrrrr} 
\toprule
 & LR & CNN & RETAIN & GRU & R-MoE \\ \midrule
AUPRC & 27.9266 & 28.4364 & 30.0276 & 30.4383 & 31.6883 \\ \bottomrule
\end{tabular}
\caption{Prediction results of all models averaged across all event types} 
\label{table:overall_results}
\vspace{-6mm}
\end{table}
% (31.6883-30.4383)/30.4383*100 = 4.10

\begin{figure}[t]
\begin{center}
\centerline{\includegraphics[scale=0.45]{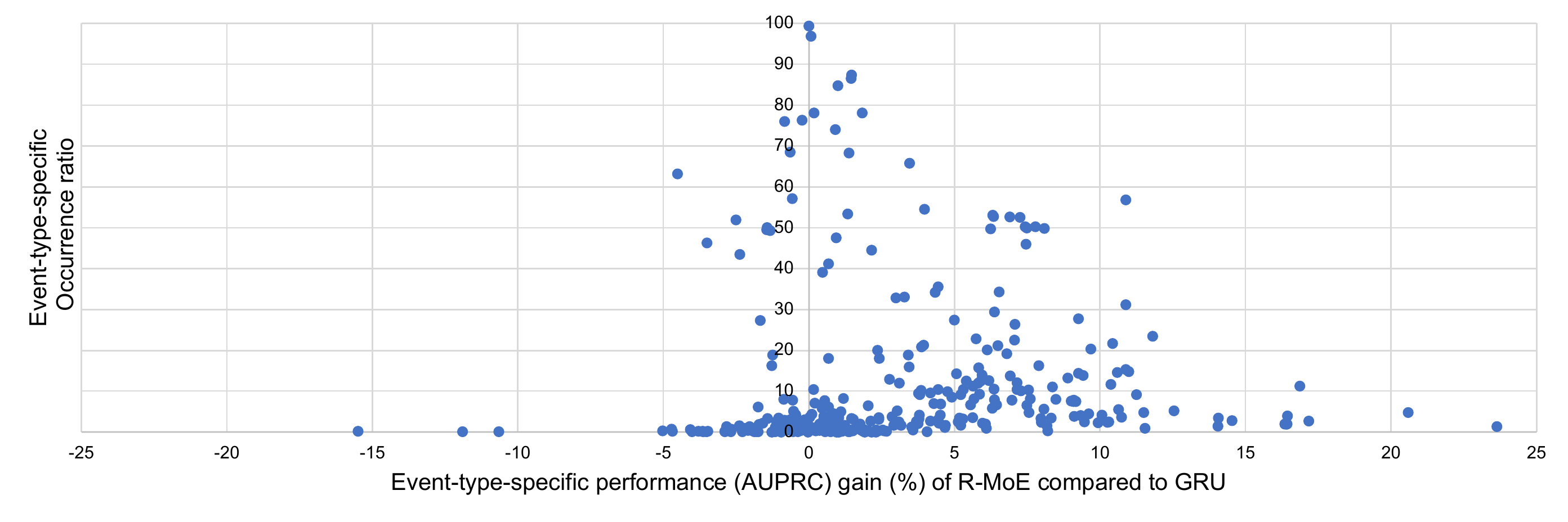}}
\caption{Event-type-specific AUPRC performance gain of R-MoE compared to base GRU model ($\delta_{base}$) and event-specific occurrence ratio. Each point represents each target event type among $E'$. Occurrence ratio is how much times each event occurred among all segmented time-windows across all test set patient data.}
\label{figure-occur-vs-auprc}
\end{center}
\end{figure}

To more understand the effectiveness of R-MoE, we look into the performance gain of our model at the individual event type level. Especially we analyze the performance gain along with the individual event type's occurrence ratio, which is computed based on how many times each type of event occurred among all possible segmented time-windows across all test set patient admissions. As shown in Fig \ref{figure-occur-vs-auprc}, we observe more performance gains are among the events that less occurred. More detailed event-specific prediction results of GRU and R-MoE models and corresponding performance gains (+\%) are presented in Appendix.  

\subsubsection{Model Capacity and Performance of R-MoE.}
\begin{figure}
    \centering
    \begin{minipage}{0.475\textwidth}
        \centering
        \centerline{\hspace{-4.5mm}\includegraphics[scale=0.48]{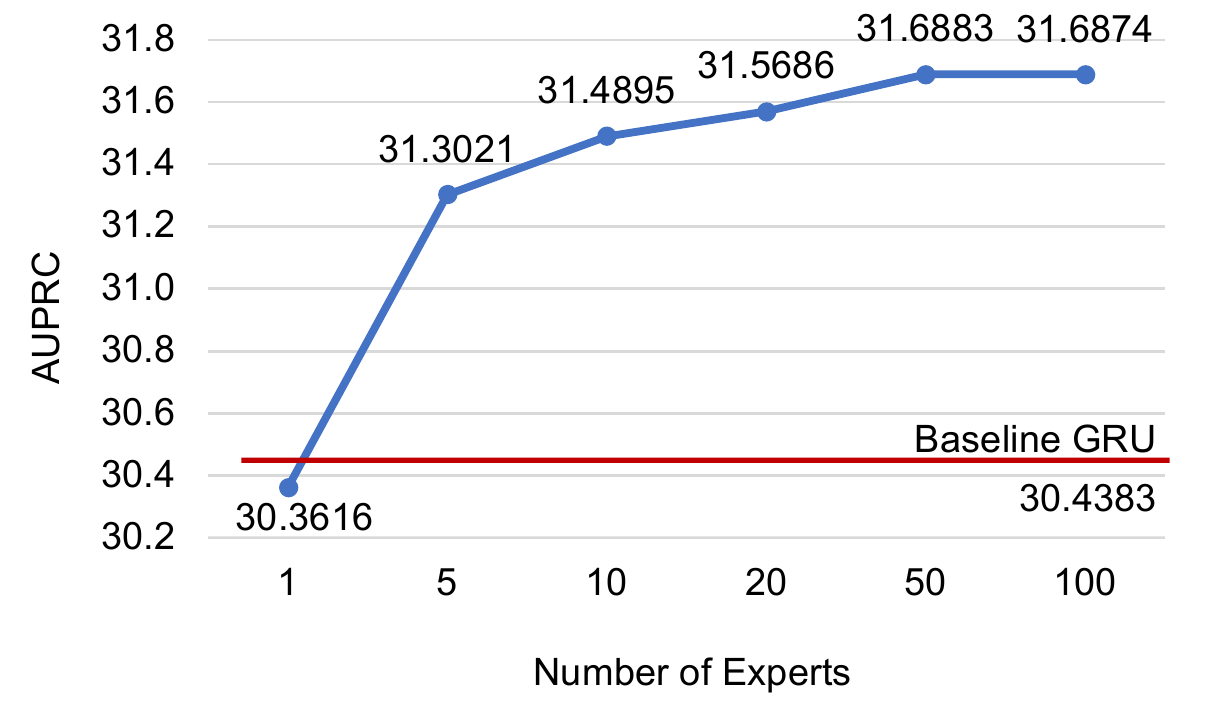}}% first figure itself
        \caption{Prediction performance of R-MoE on different number of experts. Hidden states dimension is fixed at 64.}
        \label{figure:rmoe_num_experts}
    \end{minipage}\hfill
    \begin{minipage}{0.475\textwidth}
        \centering
        \centerline{\hspace{-4.5mm}\includegraphics[scale=0.48]{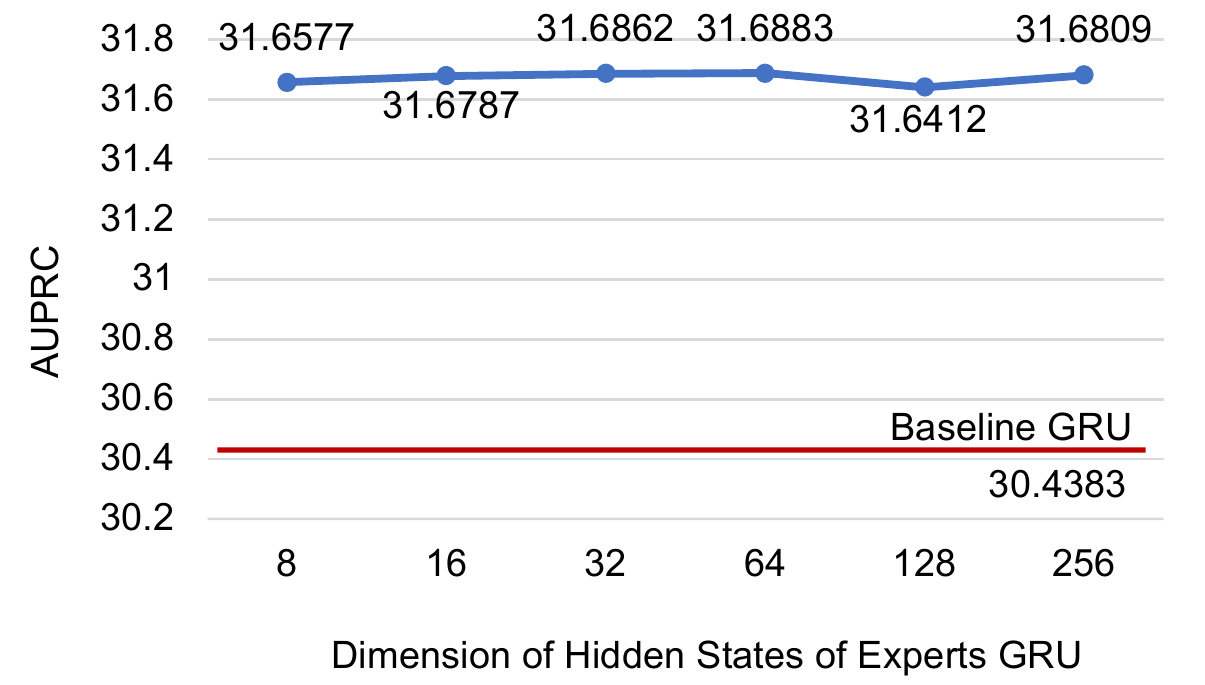}} % second figure itself
        \caption{Prediction performance of R-MoE on different hidden states dimensions. Number of experts is fixed at 50.}
        \label{figure:rmoe_hidden_dim}
    \end{minipage}
\vspace{-6mm}
\end{figure}

To further understand the performance of R-MoE regarding the various model capacities, we analyze the effect of different numbers of experts and different dimensions of hidden states on the prediction performance. Note that as written in \Cref{sec:exp-setup} the best hyperparameter is searched through internal cross-validation (number of experts = $50$ and hidden states dimension $d'$=$64$). Then, for this analysis, we fix one parameter at its best and show how the performance of R-MoE in another parameter by varying model capacity. Regarding the number of experts, a critical performance boost has occurred with a very small number of experts. As shown in Figure \ref{figure:rmoe_num_experts}, with simply five experts, we observe 2.83\% AUPRC gain compared to the baseline GRU model. With more experts, the performance is slowly increasing, but it slightly decreases after 50. Regarding different hidden states dimensions of GRU ($d'$) in R-MoE, we observe changing it does not affect much of the difference in predictive performance as shown in Figure \ref{figure:rmoe_hidden_dim}. 
% (31.3021-30.4383)/30.4383*100=2.83

\subsubsection{Ablation study.} To see the benefit of the residual learning framework in our model, we conduct an ablation study by training a simple mixture of experts (MoE) model described in Equation \ref{eq_o_moe} and comparing it with our model (R-MoE). As shown in Table 2, the residual learning framework shows its effectiveness obviously through the large margins in the evaluation metric. On average, over different model sizes, the residual learning brings 7.3\% gain in AUPRC. Notably, the simple MoE model barely outperforms the base GRU model when it has maximum experts (100).

\begin{table}[t]
\centering
\setlength{\tabcolsep}{0.5em}
\begin{tabular}{lrrrrrr} 
\toprule
Number of Experts & 1 & 5 & 10 & 20 & 50 & 100 \\ \midrule
MoE (ablation) & 28.5138 & 28.0106 & 28.4324 & 29.5318 & 30.0813 & 30.8548 \\ \midrule
R-MoE & 30.3616 & 31.3021 & 31.4895 & 31.5686 & 31.6883 & 31.6874 \\ \midrule
\% difference & 6.48 & 11.75 & 10.75 & 6.89 & 5.34 & 2.69 \\ \bottomrule
\end{tabular}
\caption{Prediction results of Mixture of Experts in Equation \ref{eq_o_moe} model (MoE) and the proposed Residual MoE model (R-MoE), differing number of experts with fixed hidden states dimension=$64$.} 
\end{table}

\section{Conclusion}

In this work, we have developed a novel learning method that can enhance the performance of predictive models of multivariate clinical event sequences, which are generated from a pool of heterogeneous patients. We address the heterogeneity issue by introducing the Residual Mixture-of-Experts model. We demonstrate the enhanced performance of the proposed model through experiments on electronic health records for intensive care unit patients.

\bibliographystyle{splncs04}
% \bibliographystyle{plain} % apalike or plain
% \begin{spacing}{0.9}
\bibliography{BibFile-Shorter}

\section*{Appendix: Event-Specific Prediction Results (AURPC)}
\vspace{-0.4cm}
\hspace{-2.5cm}\includegraphics[scale=0.80]{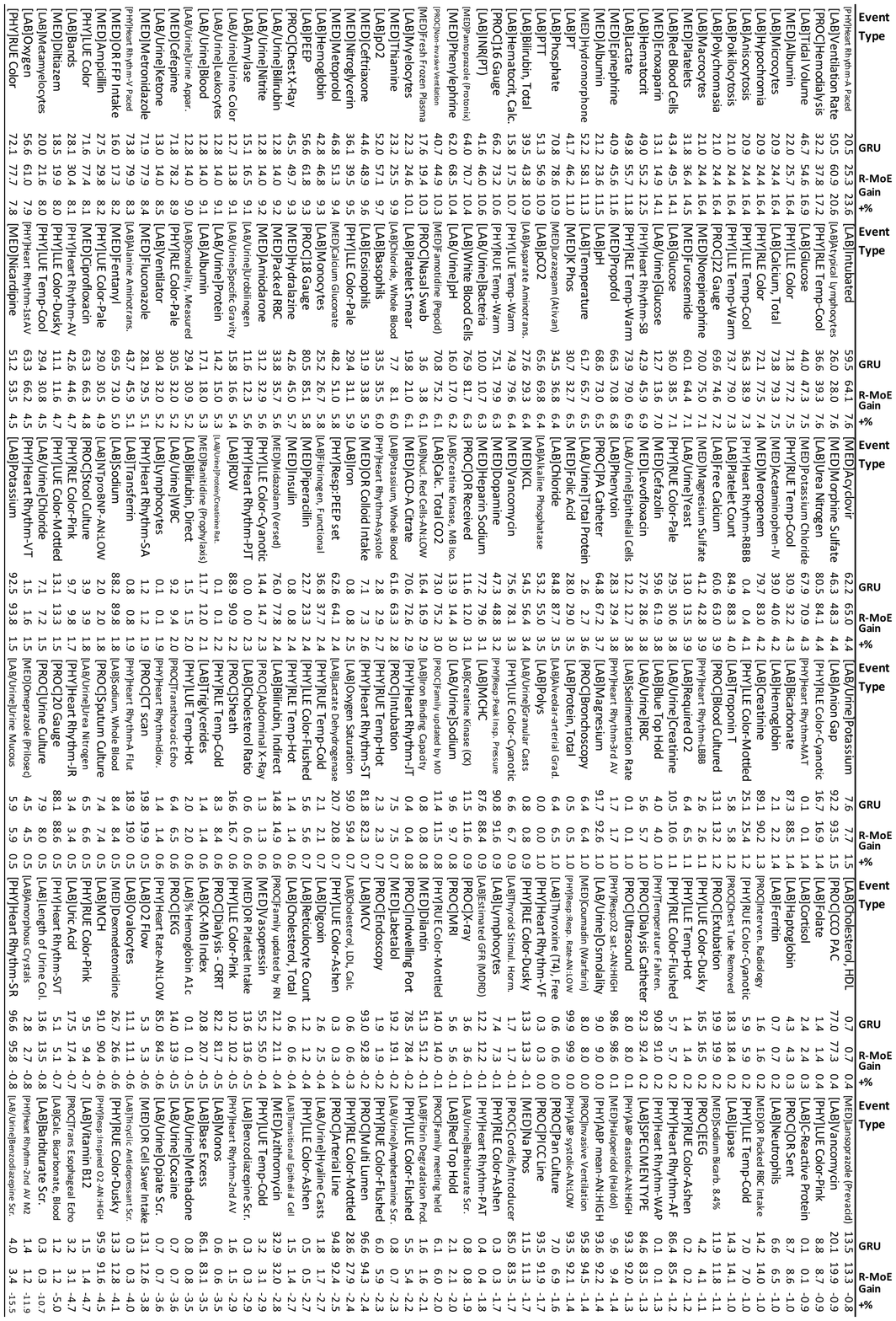}
\end{document}